\documentclass[letterpaper]{article} 
\usepackage{aaai2026}  
\usepackage{times}  
\usepackage{helvet}  
\usepackage{courier}  
\usepackage[hyphens]{url}  
\usepackage{graphicx} 
\usepackage{natbib}  
\usepackage{caption} 
\usepackage{algorithm}
\usepackage{algorithmic}
\usepackage{latexsym}
\usepackage{amssymb}
\usepackage{amsmath}
\usepackage{amsthm}
\usepackage{booktabs}
\usepackage[capitalise]{cleveref}[nameinlink=false]
\usepackage{tikz}
    \usetikzlibrary{positioning}
    \usetikzlibrary {shapes.geometric}
    \usetikzlibrary{calc}
    \usetikzlibrary{decorations.pathreplacing,intersections}

\usepackage{newfloat}
\usepackage{listings}
\DeclareCaptionStyle{ruled}{labelfont=normalfont,labelsep=colon,strut=off} 
\floatstyle{ruled}
\newfloat{listing}{tb}{lst}{}
\floatname{listing}{Listing}
\title{Advancing Safe Mechanical Ventilation Using Offline RL With Hybrid Actions and Clinically Aligned Rewards}
\author{
    Muhammad Hamza Yousuf\equalcontrib\textsuperscript{\rm 1}\thanks{Corresponding author.},
    Jason Li\equalcontrib\textsuperscript{\rm 1},
    Sahar Vahdati\textsuperscript{\rm 1,\rm 2, \rm 3},
    Raphael Theilen\textsuperscript{\rm 4},
    Jakob Wittenstein\textsuperscript{\rm 4},
    Jens Lehmann\textsuperscript{\rm 1, 5}\footnote{Work done outside of Amazon.}
}
\affiliations{
    \textsuperscript{\rm 1}Institut für Angewandte Informatik (InfAI) e. V.\\
    \textsuperscript{\rm 2}TIB - Leibniz Information Centre for Science and Technology, Hannover, Germany\\
    \textsuperscript{\rm 3}Leibniz Universität Hannover\\
    \textsuperscript{\rm 4}Department of Anesthesiology and Intensive Care Medicine, University Hospital Carl Gustav Carus, Dresden, Germany\\
    \textsuperscript{\rm 5}Amazon\\
    \{yousuf, li, vahdati\}@infai.org, \{raphael.theilen, jakob.wittenstein\}@ukdd.de, lehmann@infai.org
    
}

\begin{document}

\maketitle

\begin{abstract}
Invasive mechanical ventilation (MV) is a life-sustaining therapy commonly used in the intensive care unit (ICU) for patients with severe and acute conditions. These patients frequently rely on MV for breathing. Given the high risk of death in such cases, optimal MV settings can reduce mortality, minimize ventilator-induced lung injury, shorten ICU stays, and ease the strain on healthcare resources. However, optimizing MV settings remains a complex and error-prone process due to patient-specific variability. While Offline Reinforcement Learning (RL) shows promise for optimizing MV settings, current methods struggle with the hybrid (continuous and discrete) nature of MV settings. Discretizing continuous settings leads to exponential growth in the action space, which limits the number of optimizable settings. Converting the predictions back to continuous can cause a distribution shift, compromising safety and performance.
To address this challenge, in the IntelliLung project, we are developing an AI-based approach where we constrain the action space and employ factored action critics. This approach allows us to scale to six optimizable settings compared to 2-3 in previous studies. 
We adapt SOTA offline RL algorithms to operate directly on hybrid action spaces, avoiding the pitfalls of discretization. 
We also introduce a clinically grounded reward function based on ventilator-free days and physiological targets. Using multi-objective optimization for reward selection, we show that this leads to a more equitable consideration of all clinically relevant objectives.
Notably, we develop a system in close collaboration with healthcare professionals that is aligned with real-world clinical objectives and designed with future deployment in mind.
\end{abstract}
\begin{links}
\link{Project Website}{https://intellilung-project.eu}
\link{Code}{https://github.com/NIMI-research/intellilung-advancing-mechanical-ventilation.git}
\end{links}

\section{Introduction}
One of the most frequently applied life-saving interventions in the ICU is invasive mechanical ventilation (MV) \cite{bellani2016epidemiology}. The importance of MV was especially evident during the COVID-19 pandemic, marked by increased ICU admissions, prolonged ventilation, and early intubations.
However, MV is also associated with an increased risk of organ damage, particularly ventilator-induced lung injury (VILI) \cite{slutsky1999lung}. 
To prevent VILI, clinical guidelines recommend the limitation of tidal volumes, respiratory rate and inspiratory pressures \cite{fichtner2017s3}.
However, these protocols only provide general guidance, leaving the actual choice of ventilator setting to the clinical judgment and expertise of healthcare providers.
Furthermore, it was shown that protocols of protective MV are poorly followed worldwide \cite{bellani2016epidemiology}. MV also demands a high nurse-to-patient ratio, leading to suboptimal recovery and prolonged ICU stays in times of high workload \cite{roedl2019chronic}. 

AI-based decision support systems (AI-DSS) can provide personalized MV treatment recommendations that reduce the risk of VILI while enhancing accessibility. Offline Reinforcement Learning (RL) algorithms have demonstrated potential to learn interventions from ICU datasets that ensure immediate patient stability and improve long-term outcomes during MV \cite{kondrup2023towards,peine2021development}.

In this work, we focus on developing an AI-DSS for MV in the ICU using Offline RL. To ensure clinical relevance and practical deployability, we engage in close collaboration with clinicians and technical experts from multiple hospitals. This process shaped the identification of appropriate patient cohorts, the problem formulation, and the selection of state representations, action spaces, reward structures, and evaluation methodologies.
This paper addresses several critical technical challenges from prior studies and highlights important concerns for real-world deployment. Our contributions are as follows:

\paragraph{C1.} \label{contrib:C1}
We introduce a reward based on ventilator-free days (VFD) combined with physiological parameter safe ranges, which better aligns with the goal of reducing VILI. Previous work focused on optimizing MV using sparse rewards based on mortality. However, prior medical studies \cite{torrini2021prediction,silva2015mechanisms} indicate that mortality alone can be a poor endpoint for evaluating MV interventions. We also show how multi-objective optimization can select a reward function that balances the contributions of different clinically relevant objectives.

\paragraph{C2.} \label{contrib:C2}
Previous studies often restrict the number of MV settings because the discrete-action space grows exponentially with the number of settings. We show a simple approach to reduce the action space and adopt an additive factored critic \cite{tang2023leveragingfactoredactionspaces}.
This explicitly removes dangerous actions and reduces critic variance improving safety and performance.

\paragraph{C3.} \label{contrib:C3}
MV has both continuous and discrete settings. To avoid the pitfalls of discretizing continuous settings, we demonstrate how to adapt SOTA offline RL algorithms, namely Implicit Q-Learning (IQL) and Ensemble-Diversified Actor-Critic (EDAC), to handle (continuous and discrete) action spaces natively.

\paragraph{C4.} \label{contrib:C4}
Previous methods simplify continuous actions by discretizing them. 
However, during inference, these discrete outputs are converted back into continuous values using fixed rules or by clinicians selecting a value from the predicted bin based on their expertise. 
Our experiments show that this reconstruction introduces a distribution shift, potentially causing the learned policy to operate in regions where predictions are highly uncertain.

\section{Related Work}
In previous studies involving MV and RL, some have focused on binary decisions, such as whether to initiate MV \cite{lee2023development}, while others have addressed complex tasks including sedation and weaning strategies \cite{prasad2017reinforcement,yu2020supervised} and optimizing MV for ICU patients \cite{peine2021development,kondrup2023towards,liu2024reinforcement,roggeveen2024reinforcement,DBLP:journals/eaai/ZhangQT24,eghbali2024distributionfreeuncertaintyquantificationmechanical}.

Previous work on optimizing MV settings using offline RL either discretize the actions or purely use continuous actions. Studies that consider discretization \cite{peine2021development,kondrup2023towards,liu2024reinforcement,roggeveen2024reinforcement,eghbali2024distributionfreeuncertaintyquantificationmechanical} restrict the number of settings due to the exponential growth in action space, while studies based on continuous actions \cite{DBLP:journals/eaai/ZhangQT24} omit categorical actions. For discrete-action spaces, we address the limitations by constraining the action space to the dataset distribution and employing a factored action space. Despite these optimizations, mapping discretized actions back to continuous induces distribution shift and can produce unsafe policies, an issue largely unaddressed in prior work.
We introduce hybrid actions for two of the SOTA offline RL algorithms (IQL and EDAC), enabling them to address these issues while capturing the full range of MV settings. \cite{CHEN202247} also uses a hybrid action space for optimizing MV settings. However, they adapt an off-policy RL algorithm that lacks the safety regularization of offline methods, potentially leading to unsafe policies due to overestimation. Instead of modifying the Soft Actor-Critic (SAC) for hybrid actions using the Gumbel-Softmax reparameterization trick to allow gradient flow, we compute the expectation of the discrete component directly, as the exact distribution is available. This considerably reduces variance in policy updates \cite{christodoulou2019softactorcriticdiscreteaction}.

Most prior studies mentioned above that focus on optimizing MV have primarily relied on mortality-based rewards, either sparse or shaped. However, medical studies (see \cref{subsec:medical_context}) indicate that mortality alone is not a reliable indicator of MV treatment quality \cite{beigmohammadi2022mortality}. 
Instead, we adopt VFDs as a reward, reflecting that patients who spend less time on MV and avoid mortality received better care. Additionally, we add rewards to maintain MV-related physiological vitals within safe ranges and prevent complications.

A similar setup is used by \cite{kondrup2023towards}, where the Apache-II score function is employed as an intermediate reward, and augmented by a terminal reward based on mortality. However, combining an intermediate reward with a terminal reward is non-trivial. 
Depending on how the terminal reward is scaled, offline RL can end up ignoring one objective and overemphasize the other. Our setup optimally balances the different medical objectives. 

\section{Preliminaries}

\subsection{Offline RL} The offline RL problem is formulated as an MDP where $\{\mathcal{S}, \mathcal{A}, P, R, \gamma\}$ represents the state space, action space, transition distribution, reward distribution, and discount factor, respectively. The initial state is sampled as $s_0 \sim d_0$. 

An RL agent is trained by optimizing a policy $\pi(a|s): \mathcal{S} \rightarrow \Delta(\mathcal{A})$, guided by Q-values defined as $Q^{\pi}(s,a) = \mathbb{E}\left[\sum_{t=0}^{\infty} \gamma^t R(\cdot|s_t,a_t)\right]$. Value-based RL algorithms use the Bellman error \cite{sutton2018reinforcement} to update the Q-function. Offline RL learns from a fixed dataset $\mathcal{D}$ collected by a behavior (clinician) policy $\pi_b(a| s)$ and can suffer from overestimation due to out-of-distribution (OOD) actions. Therefore, a regularization term is often added to the standard Bellman error to mitigate this overestimation, e.g.:

\paragraph{Conservative Q-Learning (CQL).} The CQL loss function regularizes Q-values on unseen state-action pairs alongside the standard Bellman objective \cite{kumar2020conservative}. 

\paragraph{Implicit Q-Learning (IQL).} The IQL objective \cite{kostrikov2021offline} learns high-performing actions by computing action advantages via the expectile of the state value function, thereby updating the policy without querying Q-values for unseen actions.

\paragraph{Ensemble Diversified Actor Critic (EDAC).} The EDAC \cite{an2021uncertainty} uses an ensemble of critics ($Q^\pi$) to estimate the uncertainty of a given state-action pair, effectively lower bounding Q-values for uncertain pairs to prevent overestimation. Additionally, it incorporates a diversity loss among ensemble members to promote varied Q-value estimates, improving uncertainty estimation.

\subsection{MDP Formulation}

\paragraph{State.} The state space consist of 26 clinician-defined variables (see supplementary material C). These include vital signs, respiratory parameters, laboratory values, fluid balance measures, and demographic characteristics.

\paragraph{Actions.}
We consider six MV settings (control variables) listed in \cref{table:action_space}. These settings include both continuous and discrete variables. For the discrete-actions setup, all the continuous settings are discretized using clinician-defined bins. Thresholds were selected based on clinical guidelines, prior studies, and observed practice ranges, balancing clinical resolution with device limits and safety considerations (see supplementary Material~D) and aligning with the literature e.g. \cite{Grasselli2023}. The hybrid actions setup used the settings in their native continuous and categorical forms. 

We define a \emph{setpoint} as the value selected for a ventilator setting. An \emph{action} is the tuple of setpoints across all mechanical-ventilation (MV) settings,
\(a=(a_1,\dots,a_6)\), and the action space is the set of all such tuples,
$
\mathcal{A}=\mathcal{A}_1\times\cdots\times\mathcal{A}_6 .
$
In our discrete setup with clinician-defined bins,
$
|\mathcal{A}_{\mathrm{disc}}|=18{,}144.
$

\begin{table}[tb]

    \centering
    \begin{tabular}{lrr}
        \toprule
        MV Setting & Unit & Setpoint domain \\
        \midrule
        Ventilation Control Mode & - & $\{VCV,PCV\}$ \\
        Respiratory rate (RR)  & $\text{min}^{-1}$   & $[5, 60]$      \\
            Tidal Volume ($V_T$)  & $\text{ml/kg}$      & $[3, 12]$      \\
            Driving Pressure ($\Delta P$)  & $\text{cmH}_2\text{O}$ & $[0, 26]$      \\
            PEEP & $\text{cmH}_2\text{O}$ & $[0, 20]$      \\
            FiO$_2$  & $\%$                & $[21, 100]$    \\

        \bottomrule
    \end{tabular}
    \caption{Controllable MV Settings, their units and possible setpoints}
    
    \label{table:action_space}
\end{table}

\paragraph{Rewards.} Environment rewards are calculated as defined in \cref{sec:reward_design}.

\section{Clinically-Guided Reward Design (C1)}
\label{subsec:post_extub}
\label{subsec:medical_context}

Prior work (e.g., \cite{kondrup2023towards}) often uses mortality as the endpoint for evaluating MV treatment. However, mortality following a treatment is a highly confounded outcome and can be influenced by various factors, including the underlying disease and comorbidities. Instead, we worked with clinicians to define two main reward objectives aimed at improving survival, reducing the risk of VILI, and lowering healthcare costs.

\paragraph{Primary Objective.}
\label{subsec:primary_obj}
The primary objective is to reduce the duration of MV while keeping the patient alive. Prolonged MV increases the risk of complications such as ventilator-induced lung injury, infection \cite{wu2019risk, Slutsky2013}, hypotension, and diaphragm dysfunction due to disuse atrophy \cite{penuelas2019ventilator}. These complications can hinder successful weaning and are associated with increased mortality. Moreover, effective MV strategies may provide significant clinical benefit even in the absence of mortality reduction, if they facilitate earlier liberation from the ventilator \cite{Schoenfeld2002-xm}. Ventilator-free days within the first month after the start of MV are often used to assess the quality of MV. This measure combines mortality within the first month with the duration of MV, and is directly linked to the quality of ventilator settings \cite{Schoenfeld2002-xm}.

\paragraph{Secondary Objective.}
\label{subsec:secondary_obj}
The secondary objective is to limit physiological impairments due to MV. Oxygenation levels (e.g., SpO$_2$, PaO$_2$) and vital signs (e.g., blood pH, mean arterial pressure (MAP), heart rate) must remain within safe ranges to prevent adverse outcomes. For example PaO$_2$ demonstrated a U-shaped association with mortality \cite{boyle2021hyperoxaemia}, while dangerously in- or decreased blood pH-values are closely linked to organ failure and increased mortality \cite{kraut2001approach}. In collaboration with physicians, we have identified blood pH, mean arterial pressure, PaO$_2$, SaO$_2$, PaCO$_2$, heart rate and peripheral oxygen as key physiological parameters and their safe ranges to guide decision-making. These parameters were selected by clinicians from multiple hospitals within the IntelliLung consortium through a Delphi-like consensus process, ensuring broad expert agreement rather than ad-hoc tuning. Details on the parameters, their safe ranges, and reward weights are listed in supplementary E.

\subsection{Reward Design}
\label{sec:reward_design}
The total reward at each step is the sum of range reward $r_{range}$, time penalty $r_{tp}$: $r = r_{range} + r_{tp} + r_{vfd}$, and VFD reward $r_{vfd}$:

\paragraph{Range Reward $r_{range}$} guides the agent toward learning the secondary objective. 
It is calculated as follows:
\begin{equation}
    r_{range} = \frac{\sum_{i=1}^{N} w_i \cdot  \mathbf{1}_{[a_i, b_i]}(p_i) }{\sum_{i=1}^{N} w_i}, \quad r_{range} \in [0, 1]
\end{equation}

where $N$ is the total number of physiological parameters, $p_i$ is the value of the $i$-th parameter, $w_i$ is its assigned weight and $\mathbf{1}_{[a_i, b_i]}(p_i)$ is the indicator function that activates when $p_i$ is within its defined safe range $[a_i,b_i]$. 

\paragraph{Time Penalty $r_{tp}$} encourages the policy to prefer actions that shorten time on MV and to not prolong it by simply collecting positive $r_{range}$ step rewards. We set the time penalty to the negative of the maximum per‑step range reward, i.e. $r_{tp} = -1$. 

\paragraph{Ventilator Free Days}
(VFDs) \cite{Schoenfeld2002-xm} are commonly used in clinical trials to evaluate MV interventions. 
We can include VFDs in the reward function to check how well MV met the primary objective during an episode.
VFDs are calculated as follows:

\begin{equation}
   VFD =alive\:\times\max\!\bigl(0,\min(\Delta t_{re}, T_{max})-\Delta t_{mv}\bigr)
   \label{eq:vfd_eq}
\end{equation}

where $\Delta t_{re}$ is the re-intubation time, $T_{\max}$ the fixed observation window (typically around 30 days) and $\Delta t_{mv}$ the days on the ventilator.

Effectively, VFDs are the number of days a patient is both alive and free from
MV within $T_{\max}$. If the patient dies, then $\mathrm{VFD}=0$. If the patient survives, the VFDs equal the number of days spent off the ventilator, $T_{\max}-\Delta t_{mv}$. However, if the patient is re-intubated before the end
of the observation period $T_{\max}$, the VFDs are reduced to $\Delta t_{re}-\Delta t_{mv}$ \cite{Schoenfeld2002-xm}. The final VFD reward $r_{vfd}$ is:

\begin{equation}
    r_{vfd} = w_{vfd} \cdot \frac{VFD}{T_{\max}}
\end{equation}

where  $w_{vfd}$ is a hyper-parameter that controls the contribution of $r_{vfd}$. 

$r_{vfd}$ can be applied for each episode in two ways: (i) we can apply $r_{vfd}$ at the terminal time step and 0 otherwise ($VFD@TerminalStep$) or (ii) at each time step ($VFD@EachStep$).
Each option interacts differently with the weight $w_{vfd}$ and changes the relative contribution of the intermediate range reward $r_{range}$. We analyze the effect of $w_{vfd}$ and the two options in \cref{sec:results_and_discussion}.

\section{Discrete-Action Optimizations (C2)}
\label{sec:discrete_action_optimize}

Using clinician-defined bins for each MV setting (see supplementary material D), results in $|\mathcal{A}_{disc}| = 18,144$ distinct actions. However, this large action space introduces challenges, including increased computational complexity and Q-value overestimation for rarely observed actions. To address these issues, we introduce the following optimizations:

\subsection{Restrict Action Space}
\label{subsec:restrict_act_space}
The action space is restricted to only the clinician‑observed actions present in the dataset. Beyond efficiency, this constraint eliminates unsafe actions, as they do not appear in the dataset because clinicians avoid them in practice. For example, setting both $V_T$ and $FiO_2$ too low could cause severe hypoxia, leading to organ damage or worse.
RL algorithms cannot estimate the effect (Q-values) of actions absent from the dataset, and may overestimate and still select them at inference despite offline‑RL regularizations. By removing these unseen actions from the policy’s action space, we completely avoid the risk of choosing them. This reduces the action space to $53.6\%$ of $|\mathcal{A}_{disc}|$.

Based on clinical insights, we found further potential to optimize the action space. In volume-controlled ventilation (VCV) the control variable is $V_T$ and $\Delta P$ is masked.
In pressure-controlled ventilation (PCV) the control variable is $\Delta P$ and $V_T$ is masked. We implemented this by assigning a dedicated (null) bin for the masked setting.
This reduces the action space to just $6.9\%$ of $|\mathcal{A}_{disc}|$. However, this was not used in our experiments to maintain comparability with hybrid action algorithms.

\subsection{Linear Decomposition of the Q-function}

Even after reducing the discrete-action space, the critic must evaluate Q-values over a large number of actions. In offline settings with limited coverage, many state-action pairs are rare or unseen, which increases estimator variance and can introduce extrapolation bias. When the action is factored as \(a=(a_1,\dots,a_K)\), we model the critic additively,
$
Q(s,a)\approx\sum_{k=1}^{K} q_k(s,a_k),
$
following prior work on factored action spaces~\cite{tang2023leveragingfactoredactionspaces}. This reduces the output dimensionality from \(O(\prod_k |A_k|)\) to \(O(\sum_k |A_k|)\), enabling faster training and better sample efficiency because each gradient step updates multiple per-factor values. An additive critic introduces bias by omitting cross-factor interactions. However, in low-coverage regimes, the reduction in estimator variance often outweighs this bias, yielding a more favorable bias–variance trade-off. In practice, this leads to improved policies, as shown in \cref{sec:results_and_discussion}. Section H of the supplementary materials provides an illustration.

\section{Hybrid Action Space for Offline RL (C3)}
The default implementations of IQL and EDAC operate in continuous action space. We modified their CORL implementations \cite{tarasov2022corl} to support hybrid actions as follows:

\paragraph{IQL.}
For IQL, the critic function stays the same except that both continuous and one-hot encoded discrete-actions are input into the network. IQL uses Advantage-Weighted Regression (AWR) \cite{peng2019advantageweightedregressionsimplescalable} for policy optimization. The adapted $\log \pi_\phi(a|s)$ for AWR is calculated as  $\log \pi_\phi(a|s) = \log \pi_\phi^d(a^d|s) + \log \pi_\phi^c(a^c|s)$ and $(a^c,a^d)\sim\mathcal{D}$.

\paragraph{EDAC.}
EDAC is a combination of SAC \cite{haarnoja2018softactorcriticoffpolicymaximum} with ensemble of critics and a diversity loss. We follow the approach described in \cite{delalleau2019discretecontinuousactionrepresentation} for SAC adaptation, but adapted the critic to accept both discrete and continuous actions as inputs rather than outputting Q-values for each discrete-action combination. We opted for this design because empirical results showed reduced critic-loss variance and more stable training. It is important to note that the diversity loss requires $\Delta_a Q(s,a)$, which in the hybrid case becomes $\Delta_{(a^c,a^d)} Q(s,a^c,a^d)$.

\section{Experimental Setup}

\subsection{Study Datasets}
For our experiments, we use data from three publicly available clinical databases on PhysioNet \cite{goldberger2000physiobank}: MIMIC IV \cite{johnson2020mimic}, eICU \cite{pollard2018eicu} and HiRID \cite{faltys2021hirid}. The datasets include patients from different hospitals across Europe and US, ensuring broad representativeness of patient characteristics and treatment regimes. The experiment cohort includes patients aged 18 years or older who underwent at least 4 hours of MV in the ICU. Identical pre-processing steps were applied to each database. These steps included data cleaning, filtering, episode construction with 1 hour time windows, and computation \& imputation. Detailed descriptions of each step are provided in supplementary materials A.

The resulting datasets for each database (see supplementary materials B) were then combined, forming a final dataset containing 12,572 patients and 1,252,505 hours of MV. The dataset was split into 80\% training and 20\% testing. Stratified splitting was performed based on episode length and mortality while ensuring that no patient appeared in both the training and test sets. To ensure comparability, the data splits remained unchanged across all experiments.
\subsection{RL Training} 
\paragraph{Discrete-Actions.} 
We compare four CQL variants and IQL for the discrete‐action setup:
\begin{enumerate}
  \item \textbf{CQL}: Unmodified CQL as in prior work, e.g., \cite{kondrup2023towards}.
  \item \textbf{C‑CQL}: Constrained‐Action Space CQL.
  \item \textbf{F-CQL}: Factored Critic CQL.
  \item \textbf{CF‑CQL}: Constrained‑Action Space with a Factored Critic CQL(proposed discrete variant).
  \item \textbf{DiscreteIQL}: IQL adapted for discrete-action space.
\end{enumerate}
All setups use an MLP with four hidden layers of 256 units and a discount factor $\gamma = 0.99$. We set the CQL regularization to $\alpha = 0.1$ and the IQL parameters to $\tau = 0.8$ and $\beta = 5$. Gradients are clipped by global norm to 1.0, and we use a learning rate of $10^{-6}$ for CQL and $5e^{-5}$ for IQL. Each model is trained for 100,000 gradient steps, with the target critic updated after every step via Polyak averaging (coefficient $\tau_{polyak} = 0.005$).

\paragraph{Hybrid Actions.} 
We evaluate two algorithms for the hybrid‐action setup:
\begin{enumerate}
  \item \textbf{HybridIQL}: IQL adapted for hybrid actions.  It uses learning rates of $lr=1e^{-4}$, inverse temperature $\beta = 5$, expectile $\tau = 0.8$, and a MLP with four hidden layers of 256 units.
  \item \textbf{HybridEDAC}: EDAC modified for hybrid actions. We also apply automatic entropy adjustment \cite{haarnoja2019softactorcriticalgorithmsapplications} with a target entropy of $\mathcal{H}_c=-0.3$ for continuous actions and $\mathcal{H}_d =0.3$ for discrete-actions, $lr=3e^{-5}$, diversity term $\eta=1.0$ and an ensemble of 25 critics.
\end{enumerate}

\subsection{Evaluation}
\paragraph{Fitted Q-Evaluation.} FQE \cite{le2019batch} is an Off-Policy Evaluation (OPE) method used to estimate the performance of a policy $\pi$ using previously collected dataset $\mathcal{D}$. The FQE method fits a $Q^\pi$  using $y = r + \gamma Q(s',\pi(s'))$ as target. The policy performance metric $V^{\pi}$ is defined as estimated returns of policy $\pi$ on the initial state distribution: 
\begin{equation}
    V^\pi = \mathbb{E}_{s_0 \sim d_0} \left[ Q^\pi\big(s_0, \pi(s_0)\big) \right]
    \label{eq:v_pi}
\end{equation}
Since traditional FQE only captures expected returns, the distributional FQE (DistFQE) is implemented following the Quantile Regression DQN (QR-DQN) approach \cite{dabney2018distributional}. The performance of the behavior (or clinicians) policy, $V^{\pi_b}$, is evaluated by replacing the target with $y = r + \gamma Q(s', a'),$
where $a'$ is drawn from $\mathcal{D}$, and then applying \cref{eq:v_pi}.

\paragraph{Policy Coverage.}\label{subsub:policy_diverge} 
To quantify policy coverage with respect to the behavior policy at states in $\mathcal{D}$, we first estimate the behavior policy $\pi_b$. We do so by fitting a parametric model $p_{\hat{\theta}}(a| s)$ using negative log likelihood (NLL) loss:

$\theta^*
= \arg\min_{\hat{\theta}}\; \sum_{(s,a)\in\mathcal{D}} -\log p_{\hat{\theta}}(a \mid s)$

We then score $\pi$ by the expected log-density that the fitted behavior model assigns to actions drawn from $\pi$ at dataset states:
\begin{equation}
\label{eq:policy_coverage}
d^{\pi}
= \mathbb{E}_{s \sim \hat d_{\mathcal{D}},a \sim \pi(\cdot \mid s)}\
  \big[ \log p_{\theta^*}(a \mid s) \big],
\end{equation}
where $\hat d_{\mathcal{D}}$ denotes the empirical marginal over states in $\mathcal{D}$. Smaller values of $d^{\pi}$ indicate that $\pi$ tends to select OOD actions, potentially leading to poor performance and overestimation of $V^{\pi}$.

\paragraph{Reward Evaluation Metrics.} 
To evaluate how well different reward functions align with the primary and secondary objectives, we compute two metrics. The first is the Spearman correlation between the learned Q‑values and the episode $VFD$, denoted $Corr@VFD$. The second is the Spearman correlation between the learned Q‑values and the episode‑mean $r_{range}$, denoted $Corr@RangeReward$. The Q-values are estimated from $Q^\pi$ learned using FQE.

\paragraph{Tchebycheff Scalarization for Reward Selection.} 
We frame the problem of selecting a reward function as a multi-objective optimization task over objectives:$Corr@VFD$ \& $Corr@RangeReward$. We apply a range-normalized Tchebycheff scalarization \cite{Wierzbicki1980} to convert the objective vector into a single scalar score. For each objective, we compute ideal and nadir values to normalize the scales, then evaluate each candidate's normalized distance from the ideal point. The Tchebycheff value is defined as the maximum of these distances. We select the reward function with the smallest Tchebycheff value, yielding a Pareto-efficient compromise.

\section{Results \& Discussion}
\label{sec:results_and_discussion}

\subsection{Reward Functions Analysis}
\begin{figure*}[tb]
    \centering
    \includegraphics[width=\linewidth]{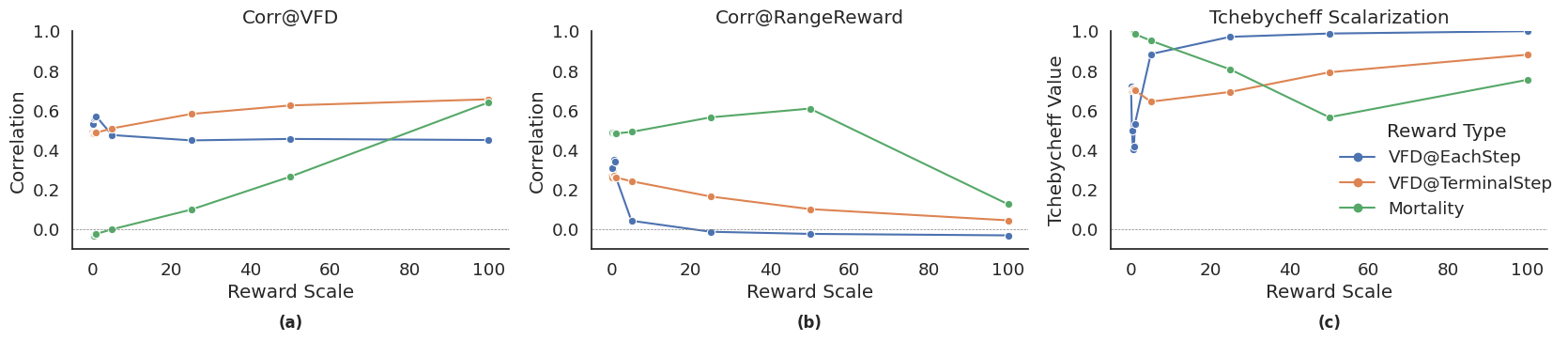}
    \caption{Panels (a)–(b) report the correlations (higher is better) \textit{Corr@VFD} and \textit{Corr@RangeReward}, respectively, across reward scales (\(w_{\mathrm{vfd}}\) for VFD and \(w_{\mathrm{morta}}\) for Mortality). Panel (c) shows the Tchebycheff value (lower is better) over the same scales. For every (reward, scale) pair, we trained five independent HybridIQL policies and report the mean of their final values.}

    \label{fig:q_reward_corr}
\end{figure*}

A high $Corr@VFD$ suggests that episodes in which the patient survives and spends fewer days on MV receive higher mean Q-values for their state–action pairs.
Similarly, a high $Corr@RangeReward$ indicates that episodes where the patient vital signs stay within safe limits are associated with higher mean Q-values.

\cref{fig:q_reward_corr} plots correlation metrics for both \textit{VFD@EachStep} and \textit{VFD@TerminalStep} across different \(w_{\mathrm{vfd}}\) values. In order to compare with mortality-based baselines \cite{kondrup2023towards,eghbali2024distributionfreeuncertaintyquantificationmechanical}, we include \(r_{\mathrm{morta}} \in [-w_{\mathrm{morta}},\, w_{\mathrm{morta}}]\) and set \(r_{\mathrm{range}}\) as the intermediate reward to directly benchmark both objectives.

\cref{fig:q_reward_corr}a shows that the time penalty $r_{tp}$ already aligns Q-values with the primary objective, as $Corr@VFD$ is already relatively high at $w_{vfd}=0$. Adding VFD strengthens this, as $Corr@VFD$ rises from $0.48$ to $0.56$ for $VFD@EachStep$ at $w_{vfd}=0.5$, and to $0.66$ for $VFD@TerminalStep$ at $w_{vfd}=100$. In contrast, alignment with the secondary objective ($Corr@RangeReward$) is highest at small $w_{vfd}$; as $w_{vfd}$ increases, intermediate rewards are effectively ignored and the correlation drops (\cref{fig:q_reward_corr}b).

Mortality-based rewards, such as those used in the prior work mentioned above, lack the time penalty and achieve a high $Corr@VFD$ only at $w_{morta}=100$; at that point $Corr@RangeReward=0.13$, meaning range rewards are largely ignored.

\paragraph{Optimal Reward Function.}
Maximizing only $\mathrm{Corr@VFD}$ or only $\mathrm{Corr@RangeReward}$ can drive the algorithm to excel on one objective at the expense of the other. Instead, we seek a Pareto‑balanced compromise. By casting reward selection as a multi-objective problem and applying range‑normalized Tchebycheff scalarization. We find that the $VFD@EachStep$ reward with $w_{vfd}\in[0.25,1]$ and in particular $w_{vfd}=0.5$ provides the best overall trade‑off (see \cref{fig:q_reward_corr}c). If the reward definition or its weighting parameters change, we rerun the same multi-objective selection to produce an updated Pareto-optimal reward function. 

\subsection{Impact of Action Discretization (C4)}
Prior work (e.g. \cite{kondrup2023towards, eghbali2024distributionfreeuncertaintyquantificationmechanical}) discretizes continuous MV settings into bins to enable training with discrete‑action offline RL algorithms. At deployment, however, the selected bin must be mapped back to a real‑valued setting. This bin‑to‑value \emph{reconstruction} can induce distribution shift, with potential performance and safety implications.

We evaluate several reconstruction functions, each mapping a chosen bin to real valued MV setting: (i) the bin mode (estimated from the dataset), (ii) a normal distribution centered at the bin mode, (iii) the bin mean, and (iv) a uniform distribution over the bin interval. We quantify the effect on policy coverage \(d^\pi\) using a policy trained with CF‑CQL. As shown in \cref{table:action_discretization_dist_mismatch}, uniform sampling yields the lowest coverage (largest distribution mismatch), whereas the bin‑mode attains the highest coverage (smallest mismatch). To ensure comparability with hybrid‑action methods that operate directly in the original action space, we therefore use bin‑mode reconstruction for discrete‑action policies when estimating \(d^\pi\).
\begin{table}[tb]

    \centering
    \begin{tabular}{lrrr}
        \toprule
        Reconstruction Method  & $d^\pi$ \\
        \midrule
 
        Bin Mode&  \textbf{-0.62}\\ 
        Gaussian Sampling & -0.70 \\ 
            Bin Mean &  -0.89\\ 
        Uniform Sampling &  -1.26\\

        \bottomrule
    \end{tabular}
    \caption{Policy coverage $d^\pi$ of a CF‑CQL policy with different reconstruction methods. }
    
    \label{table:action_discretization_dist_mismatch}
\end{table}

\subsection{Policy Analysis}
\begin{table}[tb]

    \centering
    \begin{tabular}{lrrr}
        \toprule
        Policy  & $V^\pi(\pm\sigma)$ & $d^\pi(\pm\sigma)$ \\
        \midrule
        Clinician policy &  -9.51 & -0.59 \\ 
        HybridIQL &  $-5.92\pm0.12$ &$-0.43\pm0.01$\\
        DiscreteIQL &  $-6.88\pm0.07$ &$-0.40\pm0.01$\\
        HybridEDAC & $-3.03\pm1.08$ & $-1.80\pm0.07$\\ 
        CQL &  $-5.30\pm0.24$ & $-1.84\pm0.04$\\
        CQL($\alpha=0.5$) &  $-8.11\pm0.58$ & $-1.05\pm0.19$\\
        C-CQL &  $-5.26\pm0.17$ & $-1.81\pm0.09$\\
        F-CQL &  $-6.99\pm0.19$ & $-0.52\pm0.01$\\
        CF-CQL &  $-6.82\pm0.38$ & $-0.55\pm0.01$\\ 
        \bottomrule
    \end{tabular}
    \caption{Initial-state return ($V^\pi$) and policy coverage ($d^\pi$) for policies trained using discrete- and hybrid-action algorithms, reported relative to the clinician policy. Results are reported as $\mu \pm \sigma$ over five runs with random initializations.}
    
    \label{table:algo_qvalues}
\end{table}
\begin{figure}[tb]
    \centering
    \includegraphics[width=\linewidth]{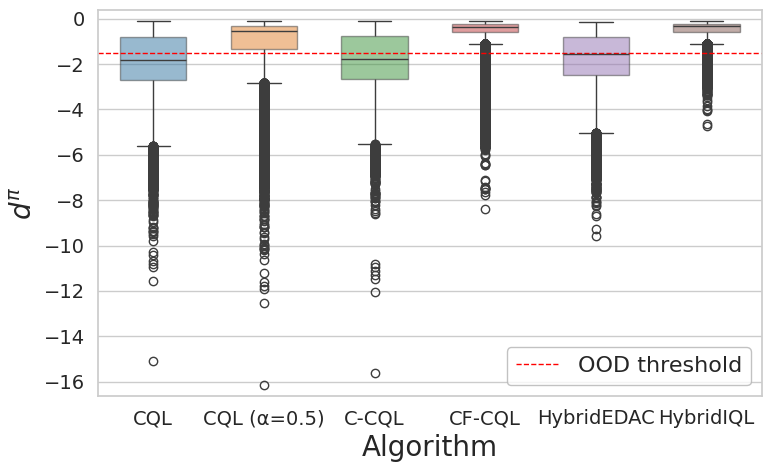}
    \caption{Distribution of policy coverage $d^\pi$ (across states in test set) for each algorithm. The red dashed line represents the OOD threshold, defined as lower
Tukey fence (at least 75\% samples lie above it) of $d^\pi$ distribution under the clinician policy. We classify actions with $d^\pi$ below threshold as OOD.}
    \label{fig:different_policy_ood}
\end{figure}

\begin{table}[tb]

    \centering
    \begin{tabular}{lrrr}
        \toprule
        Policy  & $V^\pi(\pm\sigma)$ & $d^\pi(\pm\sigma)$ \\
        \midrule
        Clinician policy &  -12.50 & -0.74 \\ 
        HybridIQL &  $-6.40\pm0.18$ &$-0.55\pm0.01$\\
        HybridEDAC & $-2.56\pm1.23$ & $ -1.65\pm0.10$\\ 
        CQL &  $-6.85\pm0.88$ & $-1.46\pm0.10$\\
        
        CF-CQL &  $-7.57\pm0.21$ & $-0.80\pm0.05$\\ 
        \bottomrule
    \end{tabular}
    \caption{Initial-state return ($V^\pi$) and policy coverage ($d^\pi$), trained on eICU and HiRID and evaluated on MIMIC-IV for external generalizability, reported as $\mu \pm \sigma$ over five random initializations.}
    
    \label{table:external_eval}
\end{table}

\begin{table}[tb]

    \centering
    \begin{tabular}{lrrr}
        \toprule
        Policy  & $V^\pi(\pm\sigma)$ & $d^\pi(\pm\sigma)$ \\
        \midrule
        Clinician policy &  -9.51 & -0.59 \\ 
        HybridIQL &  $-6.19\pm1.23$ &$-0.38\pm0.04$\\
        HybridEDAC & $-7.14\pm3.63$ & $-1.61\pm0.54$\\ 
        CQL &  $-8.16\pm1.15$ & $-0.91\pm0.37$\\
        
        CF-CQL &  $-9.11\pm1.99$ & $-0.62\pm0.30$\\ 
        \bottomrule
    \end{tabular}
    \caption{Initial-state return ($V^\pi$) and policy coverage ($d^\pi$), reported as $\mu \pm \sigma$ over a hyperparameter sweep (learning rate, parameter count, offline RL regularization) and two random initializations. We report results from 162 runs for each of HybridIQL and HybridEDAC, and 54 runs for each of CQL and CF-CQL.
}
    
    \label{table:robustness_to_hyper}
\end{table}

In this section, we evaluate the performance and reliability of discrete and hybrid-action offline RL algorithms using FQE estimates ($V^\pi$) and policy coverage ($d^\pi$).

\paragraph{Discrete-Action Optimizations.}
In \cref{table:algo_qvalues} we compare the unmodified CQL from previous work, e.g \cite{kondrup2023towards} with our proposed variant CF-CQL. While unmodified CQL achieves the highest estimated value ($V^\pi$), it suffers from very low policy coverage ($d^\pi$), indicating over-reliance on OOD actions. This is confirmed by \cref{fig:different_policy_ood}, where most CQL-selected actions lie beyond the OOD threshold.
In contrast, CF-CQL maintains most actions within the in-distribution (ID) region and achieves a more reliable value estimate, albeit a bit lower. To disentangle which component drives the gains, we also test variants with only constrained-action (C-CQL) and only factored-critic (F-CQL). C-CQL provides marginal improvement over CQL, while F-CQL performs similarly to CF-CQL, confirming that the improvement in coverage ($d^\pi$) stems from the factored critic, whereas constraining actions mainly serves to eliminate unsafe actions.

\paragraph{Higher Critic Regularization.} We also experimented with a higher CQL critic regularization ($\alpha=0.5$) to increase policy coverage. Increasing $\alpha$ leads to higher $d^\pi$, but at the cost of significantly reduced $V^\pi$, while $d^\pi$ still remained below that of CF-CQL. This suggests that regularization alone is insufficient to match both the performance and distributional robustness achieved through factored critics.

\paragraph{Hybrid-Action Algorithms.} Among hybrid-action methods, HybridEDAC achieves the highest $V^\pi$, but like CQL, its low $d^\pi$ suggests heavy reliance on OOD actions, raising concerns about potential overestimation. In contrast, HybridIQL strikes a more favorable trade-off: it achieves competitive value estimates while maintaining significantly higher policy coverage across both discrete and hybrid methods, highlighting the advantage of the hybrid-action setup.

\paragraph{Are hybrid-actions needed?} To assess whether the gains of HybridIQL are due to the hybrid action formulation or IQL itself, we also compare it with DiscreteIQL in \cref{table:algo_qvalues}. HybridIQL achieves higher $V^\pi$ while maintaining the most favorable trade-off between performance and distribution coverage $d^\pi$. As discussed previously, the discrete-action setups are also affected by distribution shifts at inference time due to bin-to-value reconstruction, which may pose additional safety concerns.

\paragraph{Generalization.} To test cross-dataset generalization, we train only on eICU and HiRID and test on MIMIC-IV. Results are given in \cref{table:external_eval} and closely follow \cref{table:algo_qvalues}. HybridIQL shows the best generalization by achieving highest performance $V^\pi$ while having  policy coverage $d^\pi$ higher than the clinician policy.

\paragraph{Robustness to hyper-parameters.}
\cref{table:robustness_to_hyper} reports a sweep over learning rates 
$\{10^{-4}, 10^{-5}, 10^{-6}\}$, network capacities 
$\{[64, 64], [128, 128, 128], [256, 256, 256]\}$, 
offline RL regularization coefficients for CQL variants 
$\alpha \in \{0.1, 0.5, 2.5\}$, for HybridIQL 
$\tau \in \{0.7, 0.8, 0.9\}$ and $\beta \in \{5, 10, 25\}$, 
for HybridEDAC $\eta \in \{0.1, 1, 10\}$ and total number of critics 
$\in \{5, 10, 25\}$, as well as two random seeds. 
HybridIQL is the most stable, having high coverage and strong initial-state value $V^\pi$ across the sweep, followed by CF-CQL. By contrast, HybridEDAC and CQL exhibit higher variance and consistently poorer coverage.

\begin{figure}[tb]
    \centering
    \includegraphics[trim = 1cm 1cm 1cm 0cm, width=0.9\linewidth]{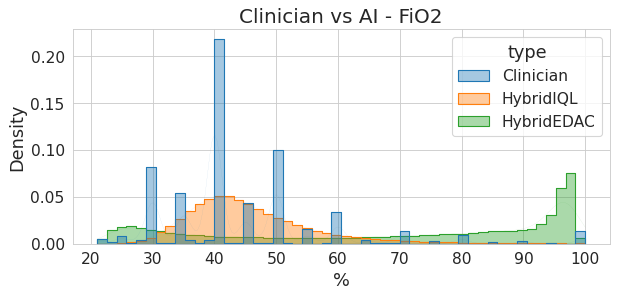}
    \caption{Action distribution of FiO$_2$ for hybrid-action setup vs clinician.}
    \label{fig:vt_distribution}
\end{figure}

\begin{figure}[tb]
    \centering
    \includegraphics[trim = 1cm 1cm 1cm 0cm, width=0.9\linewidth]{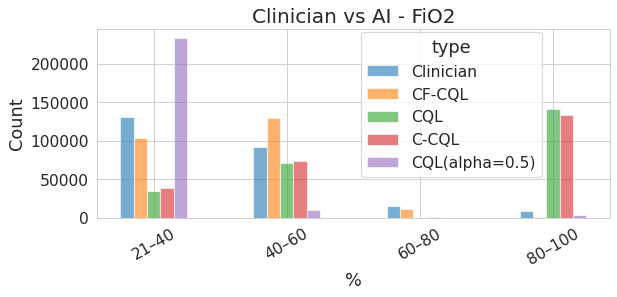}
    \caption{Action distribution of FiO$_2$ for discrete-action setup vs clinician.}
    \label{fig:vt_distribution_disc}
\end{figure}

\paragraph{Action Distribution.}\cref{fig:vt_distribution} presents the policy distribution of $\mathrm{FiO_2}$ in the hybrid setting, and \cref{fig:vt_distribution_disc} shows the discrete setting.  Clinician baseline puts little probability mass above $80\%$. However, the low $d^\pi$ policies (HybridEDAC, CQL, C-CQL) overestimate in these low-density regions (high variance). In contrast, CQL with a higher $\alpha$ ($0.5$) is overly conservative, placing most mass where the clinician density is highest and leading to a lower $V^\pi$ (high bias). Algorithms with higher $d^\pi$ (CF-CQL, HybridIQL) better balance the bias–variance trade-off and achieve competitive performance.
A complete set of action-distributions plot is provided in the supplementary material F.

\paragraph{Key Takeaways.}Focusing solely on high performance ($V^\pi$) is insufficient for safe deployment when a policy frequently selects OOD actions (i.e., has low $d^\pi$). Prioritizing algorithms that minimize distributional shift yields more trustworthy performance and increases the likelihood of real-world adoption. 
We find that our discrete-action optimizations remove unsafe actions and improve distributional coverage, enabling CQL to scale to six actions (vs. two–three in prior work). In parallel, the hybrid-action method, specifically HybridIQL, avoids inference-time distribution shift inherent to discretization and delivers higher performance, better coverage, stronger cross-dataset generalization, and greater hyperparameter robustness than discrete baselines. Taken together, these results indicate that enforcing coverage and leveraging hybrid actions provide a principled path toward AI-DSS that reduce ventilator-induced lung injuries and enable safer mechanical ventilation.


\subsection{Limitations \& Future Research} 
While offline RL methods inherently avoid OOD actions, the agent may still propose unsafe actions if such actions are present in the dataset. In future work, we aim to address this by collaborating with clinicians to identify and avoid potentially unsafe action patterns by incorporating national and international ventilation guidelines \cite{Grasselli2023,Qadir2024}.
Furthermore, our study relies on publicly available ICU datasets that suffer from noise, limited resolution, and require assumptions about patient states and interventions. Although our datasets span multiple US and European hospitals, exceeding prior single-center or US-only studies, we acknowledge no dataset captures all ICU scenarios.
To mitigate this, we are currently collaborating with partner hospitals in Germany, Spain, Poland, Italy and the US to collect high-quality, prospectively curated data. 
This system is designed for recommendation only, as full agency and responsibility reside with the attending physician. Still, the potential of recommendations to bias clinical decision-making warrants further consideration.
Lastly, all evaluations were retrospective and offline. We used FQE, coverage, and OOD thresholding to estimate performance and provide empirical safety guarantees supporting initial retrospective claims. However, prospective validation of the observed benefits is required. Ongoing observational ICU trials are assessing real-world performance and safety without altering care, with a randomized controlled trial planned to evaluate efficacy and safety against standard care in accordance with regulatory best practices.

\section{Conclusion}
This research tackles the critical challenge of suboptimal MV settings, a persistent issue in ICUs that contributes to preventable harm and mortality worldwide. By advancing offline RL methods in this context, our work supports the development of safer, more reliable AI-DSS for MV. Specifically, we addressed key technical challenges by incorporating constrained and factored action space optimization for discretized setup to increase MV settings recommended to clinicians. 
Our work also identified that discretizing actions can introduce distribution shifts, for which we studied the impact of different reconstruction methods. In parallel, we extended SOTA offline RL algorithms (IQL \& EDAC) to incorporate hybrid actions to overcome the limitations of discretization. CF-CQL and HybridIQL demonstrated improvements over clinician policies while offering trade-offs between enhanced safety and performance. Additionally, in cooperation with clinical experts, we introduced rewards based on VFD and safe ranges, which are better aligned with clinical objectives. Overall, the methods presented in this study contribute to a more robust adaptation of offline RL in MV by paving the way for a prospective multi-hospital study in Germany, Spain and Poland scheduled to commence later this year. While our focus was on MV, the developed methodologies may also have broader applicability beyond this specific clinical task.

\section*{Ethical Statement}
We used publicly available (MIMIC IV, eICU and HiRID), de-identified data in compliance with relevant usage agreements and applicable data protection laws. This retrospective analysis did not involve any direct patient interventions or modifications in patient care.
We recognize that our reward function is specified at the cohort level and may not fully capture the needs, risks, or preferences of specific individuals or minority subgroups, which could lead to systematic harm. To mitigate this, we constrain the learned policy to remain close to clinician-like behavior, ensure that any deployment would occur only under active clinician oversight, and plan explicit subgroup- and individual-level safety analyses in future prospective studies.

\section*{Acknowledgments}
Our work is supported by the European Union project IntelliLung with Grant Agreement No. 101057434. The authors gratefully acknowledge the support of the IntelliLung Consortium \footnote{\url{https://intellilung-project.eu/consortium-members}}.

\bibliography{references}
\onecolumn
\appendix
\section*{Supplementary}

\section{Pre-processing steps}
This section describes the pre-processing steps performed on each dataset. Identical pre-processing steps were applied to each database, using individual pipelines to account for dataset-specific characteristics.

\begin{itemize}
    \item \textbf{Data cleaning} consisted of standard cleaning steps such as unit conversion and outlier removal. String values like ventilation mode and sex were encoded numerically.
    \item \textbf{Data filtering}  retained only mechanical ventilation (MV) periods meeting the minimum duration requirement of 4h. Patients fully missing any of the required variables were excluded. 
    \item \textbf{Episode building}  involved defining ventilation episodes and time steps. Episodes were identified using invasive ventilation identifiers or, when unavailable, inferred from ventilation-specific variables. A gap of at least 6 hours between ventilation variables marked the end of one episode and the start of another. This threshold was defined in collaboration with clinical partners to ensure meaningful episode segmentation, avoiding unnecessary splits for short gaps while accounting for potential clinical changes over longer gaps. For each episode, 1-hour time steps were created. When multiple values were available within a time step, a rule-based selection using Logical Observation Identifiers Names and Codes (LOINC) was applied, prioritizing measurements based on clinical relevance, such as method of collection. If unresolved, the median was chosen for numerical variables, while for categorical variables, the values with the longest duration within the time window was selected. 
    \item \textbf{Computation \& Imputation} included calculating values for the state variables (e.g., cumulative fluids intake/4h, mean arterial pressure, etc.) and reward (e.g., $\Delta t_{mv}$, $\Delta t_{re}$), as well as imputing missing data within an episode using forward propagation.
\end{itemize}

\section{Dataset statistics}
Table \ref{table:state_vector_details} provides an overview of the datasets used in the experiments, including MIMIC-IV, eICU, and HiRID. 
The table presents key statistics for each dataset, including the number of patients, the number of ventilation episodes, and the total hours of MV recorded. 
\begin{table}[htb]
    \centering
    \caption{Statistics for each dataset after pre-processing.}
    \begin{tabular}{lrrr}
        \toprule
        \textbf{Dataset} & \textbf{No. Patients} & \textbf{No. Episodes} & \textbf{Hours of MV} \\
        \midrule
        MIMIC IV & 1,538 & 1,616 & 154,296 \\
        eICU & 5,678 & 5,804 & 678,740 \\
        HiRID & 5,368 & 5,551 & 422,477 \\
        \bottomrule
    \end{tabular}
    
    \label{table:state_vector_details}
\end{table}

\section{Variable ranges}\label{appendix:variable_ranges}

Table \ref{table:variable_details} presents a comprehensive list of physiological and laboratory variables used in the experiments. 
It includes their respective measurement ranges, units, and corresponding LOINC. 
The ranges reflect observed values within the dataset, ensuring consistency in the experimental setup

\begin{table}[H]
    \centering
    \caption{Variable ranges, units and LOINC codes used for the experiments}
    \small
    \begin{tabular}{llrrr}
        \toprule
        \textbf{Group} & \textbf{Variable} & \textbf{Ranges} & \textbf{Unit} & \textbf{LOINC} \\
        \midrule
         Circulatory function & Mean Arterial Pressure & 30-200 & mmHg & 8478-0  \\ 
         & Diastolic Pressure & 20-120 & mmHg & 8500-1 \\
         & Systolic Pressure & 50-260 & mmHg & 8480-6  \\
         Ventilation & Inspiratory Airway Pressure & 10-60 & 10-60 & 75942-3 \\
         & Tidal Volume (observed) & 80-2040 & ml & 75958-9 \\
         Haematology & Haemoglobin & 2-20 & mmol/l & 718-7 \\
         & White Blood Cell Count & 0-30 & Gpt/l & 26464-8 \\
         Gas exchange & SaO2 & 40-100 & \% & 2708-6  \\
         & SpO2 & 30-100 & \% & 59408-5 \\
         & PaO2 & 20-600 & mmHg & 19255-9 \\
         & PaCO2 & 20-100 & mmHg & 32771-8 \\
         Blood gas analysis & Base excess & -20-30 & mmol/l & 11555-0 \\
         & pH & 6-8 & - & 97536-7 \\
         Fluid status & Intravenous Fluid Intake & 0-20000 & ml/4h & (multiple) \\
         & Urine Output & 0-2000 & ml/4h & (multiple) \\
         & Vasopressors & 0-5 & NE & (multiple)\\
         Electrolytes & Potassium & 2-10 & mmol/l & 75940-7 \\
         & Chloride & 80-150 & mmol/l & 2069-3 \\
         & Sodium & 120-180 & mmol/l & 2947-0 \\
         Coagulation & INR & 0.9-15 & - & 34714-6 \\
         Heart function & Heart Rate & 20-200 & 1/min & 8889-8 \\
         Demographics & Age & 18-120 & years & 30525-0 \\
         & Sex & 0-1 & - & 72143-1 \\
         & Weight & 40-140 & kg & 29463-7 \\
         & Height & 155-200 & cm & 8302-2 \\
        \bottomrule
    \end{tabular}
    
    \label{table:variable_details}
\end{table}

\section{Action bins}
\label{appendix:action_bins}
The bins in \cref{table:action_bins} were defined through a Delphi-like process involving a panel of clinical experts from the United States, Europe, and Australia. Thresholds were determined based on clinical guidelines, published studies, and observed practice ranges. The selected values balance clinical resolution with device limitations and safety considerations, consistent with prior literature, e.g. \cite{Grasselli2023}.
Bin thresholds were defined as left-inclusive. For the ventilation mode-dependent actions \textit{Driving Pressure} and \textit{PEEP}, the final bin "n.a" indicates that the setting is not controllable in that mode.

\begin{table}[H]
    \centering
    \caption{Bin indices and their ranges for action discretization}
    \begin{tabular}{lrrrrrrrrr}
        \toprule
        Action & 1 & 2 & 3 & 4 & 5 & 6 & 7 & 8 & 9 \\
        \midrule
        Ventilation control mode & 0 & 1 &  &  &  &  &  &  &  \\ 
        Respiratory rate & 5-10 & 10-15 & 15-20 & 20-25 & 25-30 & 30-35 & 35-60 &  &  \\ 
        Tidal Volume & 3-4 & 4-5 & 5-6 & 6-7 & 7-8 & 8-9 & 9-10 & 10-11 & 11-12 \\ 
        Driving Pressure & 0-6 & 6-10 & 10-14 & 14-18 & 18-22 & 22-26 & 26-40 & n.a. &  \\ 
        PEEP & 0-4 & 4-8 & 8-12 & 12-16 & 16-20 & 20-50 & n.a. &  &  \\ 
        FiO2 & 21-40 & 40-60 & 60-80 & 80-100 &  &  &  &  &  \\ 
        \bottomrule
    \end{tabular}
    
    \label{table:action_bins}
\end{table}

\section{Clinically defined rewards}
The range reward $r_{range}$ checks whether physiological parameters are within safe ranges. For this study, we consider the parameters listed in \cref{table:range_reward}. 
The parameters were co-developed with clinicians from multiple hospitals (United States, Europe, Australia) using a Delphi-like consensus process, ensuring broad expert agreement rather than ad-hoc tuning. Weights reflect clinician priorities and support training stability. The design aligns with prior literature, e.g. \cite{kraut2001approach}, but may be revisited as clinical guidelines and expert consensus evolve.

\begin{table}[H]

    \centering
    \begin{tabular}{lrr}
        \toprule
        Physiological Parameters  & Safe Range & Weight \\
        \midrule
        Blood pH & $[7.3,7.45]$ & 2  \\ 
        Mean Arterial Pressure& $[60,109]$ & 1 \\ 
        PaO$_2$ & $[55,80]$ & 2 \\ 
        SaO$_2$ & $[88,96]$ & 2 \\ 
        PaCO$_2$ & $[28,55]$ & 2 \\
        Heart Rate & $[70,109]$ & 1 \\
        Peripheral Oxygen & $[88,96]$ & 2 \\

        \bottomrule
    \end{tabular}
    \caption{Parameters considered for range reward along with their safe ranges and weights.}
    
    \label{table:range_reward}
\end{table}

\section{Action distributions}\label{appendix:full_action_distributions}

\begin{figure}[H]
    \centering
    \includegraphics[width=0.49\linewidth]{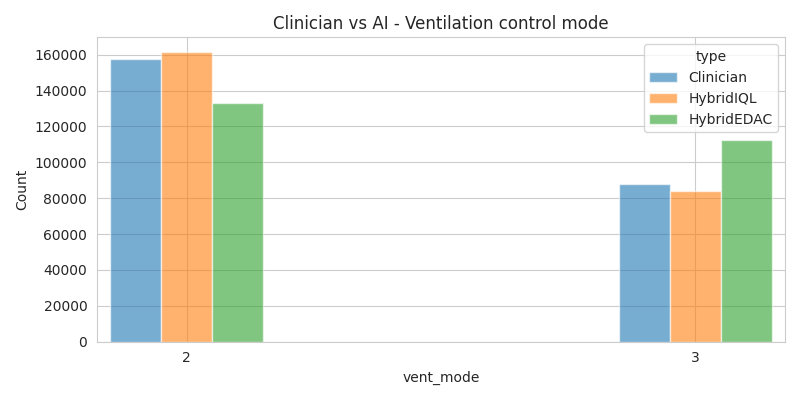}
    \includegraphics[width=0.49\linewidth]{figures/action_distributions/hybrid/vent_fio2.png}
    
    \includegraphics[width=0.49\linewidth]{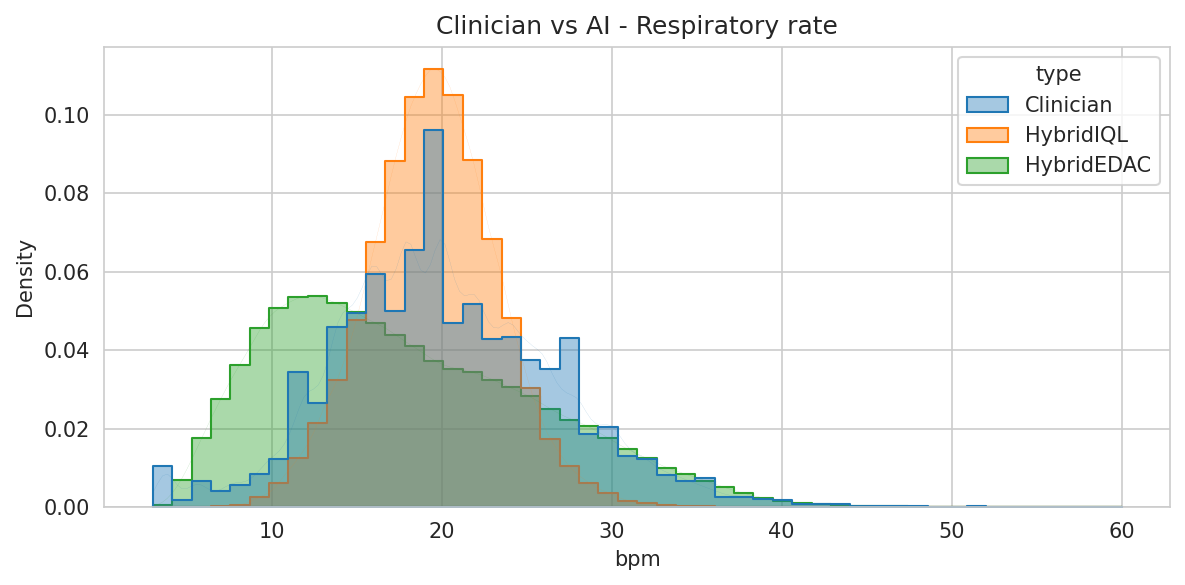}
    \includegraphics[width=0.49\linewidth]{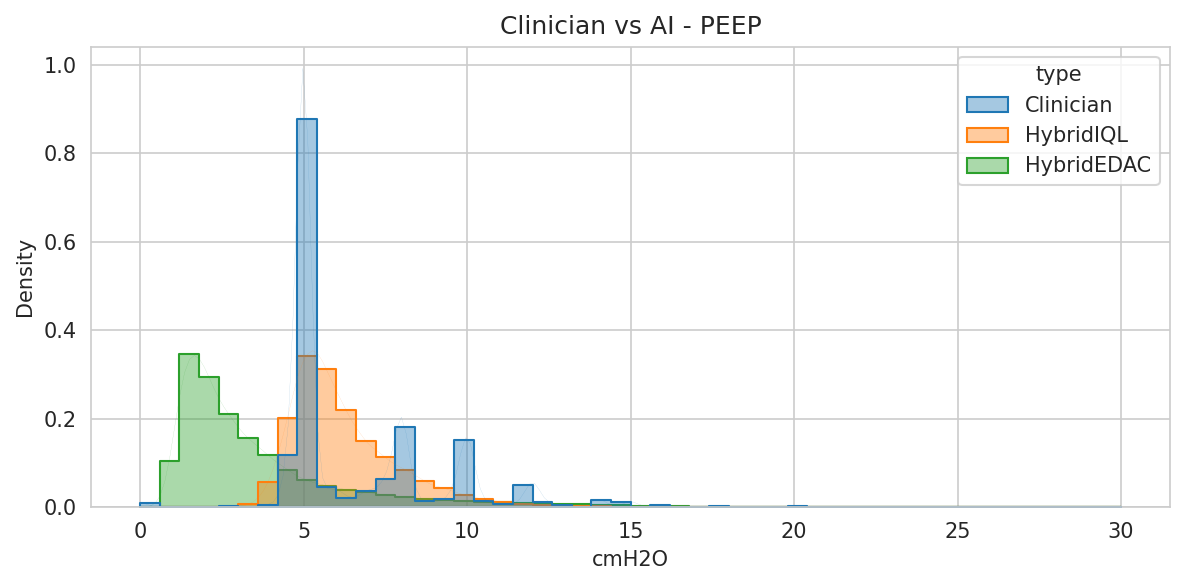}
    
    \includegraphics[width=0.49\linewidth]{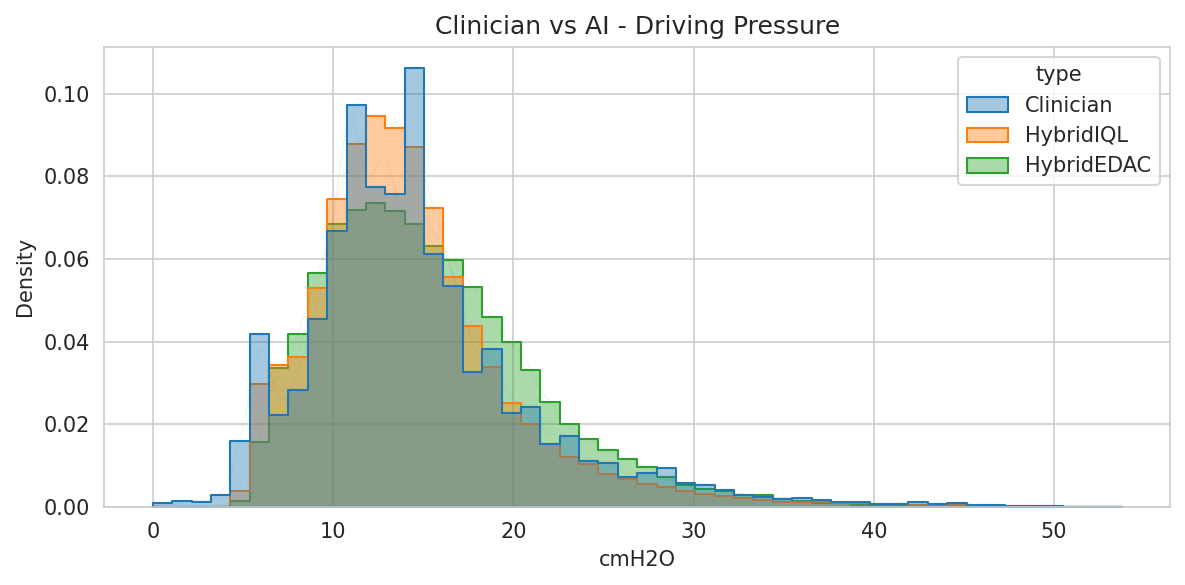}
    \includegraphics[width=0.49\linewidth]{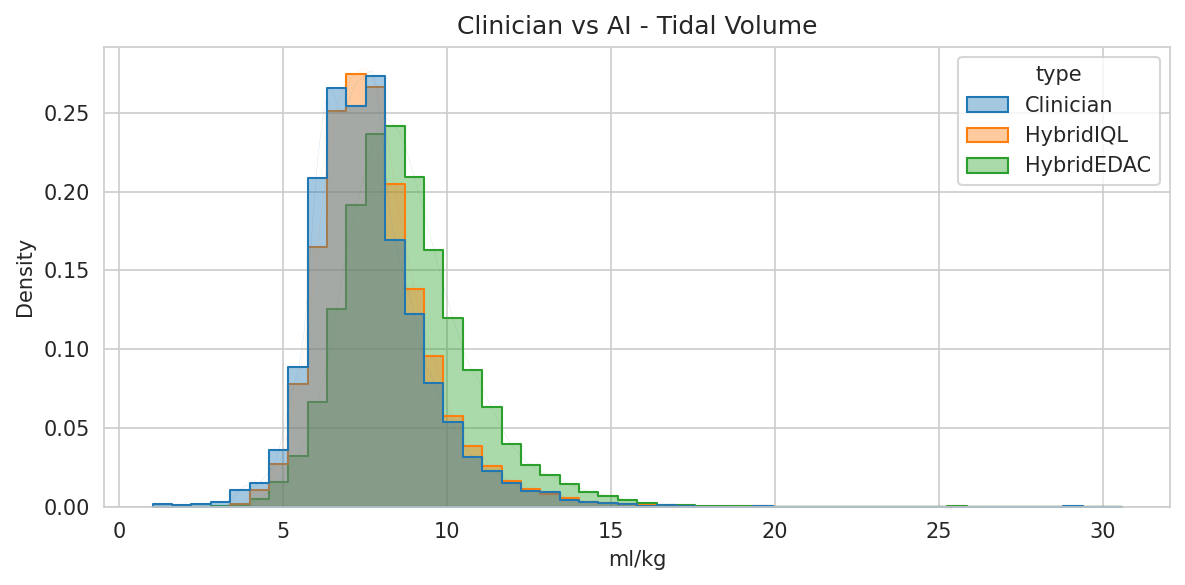}
    
    \caption{Hybrid-Action setup action distributions vs Clinician}
    \label{fig:full_action_distributions}
\end{figure}

\begin{figure}[H]
    \centering
    \includegraphics[width=0.49\linewidth]{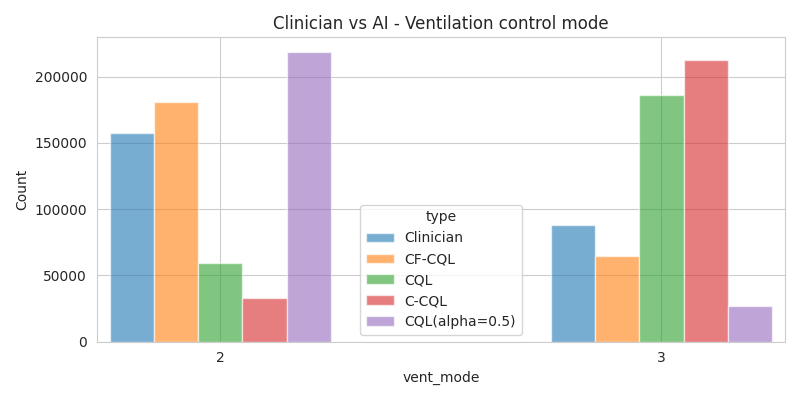}
    \includegraphics[width=0.49\linewidth]{figures/action_distributions/discrete/vent_fio2.png}
    
    \includegraphics[width=0.49\linewidth]{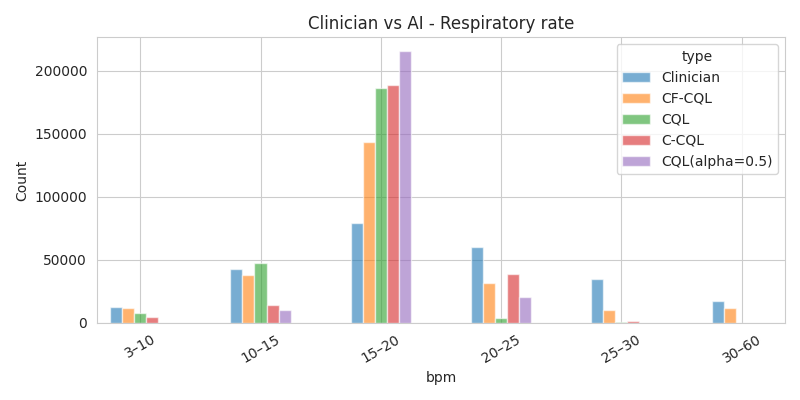}
    \includegraphics[width=0.49\linewidth]{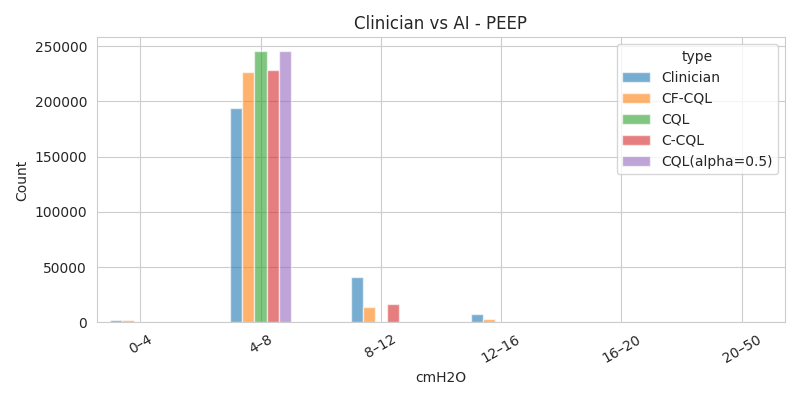}
    
    \includegraphics[width=0.35\linewidth]{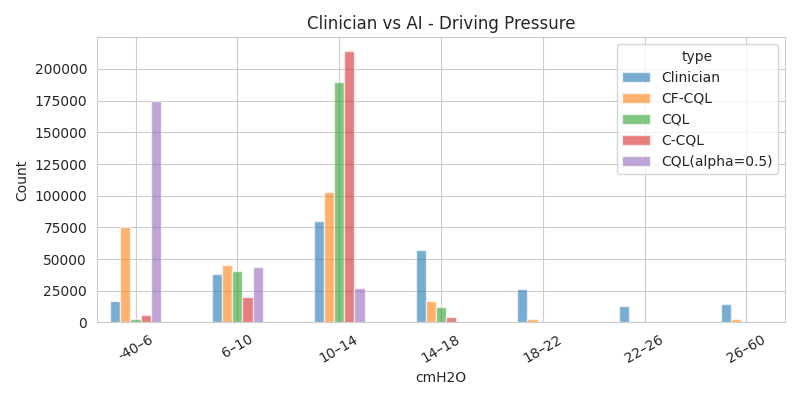}
    \includegraphics[width=0.49\linewidth]{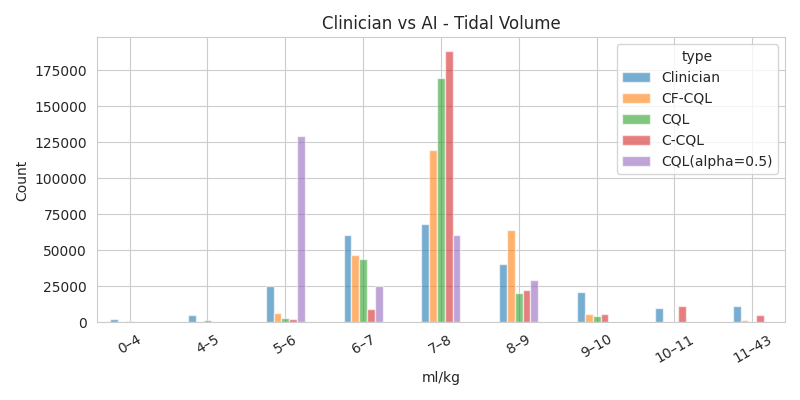}
    
    \caption{Discrete-Action setup action distributions vs Clinician}
    \label{fig:full_action_distributions_disc}
\end{figure}

\cref{fig:full_action_distributions} and \cref{fig:full_action_distributions_disc} show the distribution comparisons of different actions between clinicians and the trained policies for continuous and discrete models, respectively.

\section{Critic Factorization}
\begin{figure}[H]
   \centering
\begin{tikzpicture}[Brace/.style={
     thick,
     decoration={ 
        brace,
        amplitude=5pt 
        },
     decorate 
  },
roundnode/.style={circle, draw=black!60, fill=black!5, minimum size=5mm},
operation_rec/.style={rectangle, draw=black!60, fill=black!5, minimum width=1mm, minimum height=1mm, font=\scriptsize, align=center},
item_node_green/.style={rectangle, draw=green!60, fill=green!5, minimum width=1mm, minimum height=1mm, font=\scriptsize, align=center},
squarednodered/.style={rectangle, draw=red!60, fill=red!5, very thick, minimum size=5mm},
squarednodegreen/.style={rectangle, draw=green!60, fill=green!5, very thick, minimum size=5mm},
arrowstyle/.style={->, thick, black!70!black, shorten >=1mm, shorten <=1mm},
noarrow/.style={-, thick, black!70!black, shorten <=1mm},
textstyle/.style={font=\scriptsize, align=center}, 
rectstyle/.style={rectangle, draw=black, minimum width=1mm, minimum height=1mm, font=\scriptsize, align=center},
trapez/.style={trapezium, fill=#1!20, draw=#1!75, text=black}
]


\node [trapez=blue, minimum width=1.5cm, minimum height=0.75cm,trapezium stretches body, rotate=-90, xshift=-0.71cm] (q_net) {\rotatebox{90}{Critic}};

\node[rectstyle] (state) [left=0.5cm of q_net.south] {$S$};

\draw[arrowstyle] (state) to (q_net.south);

\node[textstyle] [right=0.5cm of q_net.north] (q_out) {$\{Q(s,a_{0,0}) \dots Q(s,a_{k,l})\}$};
\draw [Brace] (q_out.north west) -- (q_out.north east)
    node[midway, above=5pt, font=\small] {$batch\_size \times 37$};

\draw[arrowstyle] (q_net.north) to (q_out);

\coordinate (q_out_anchor) at ($(q_out) + (2,0)$);
\node[roundnode]      (cross)                       [below=0.5cm of q_out_anchor]   {$\times$};

\draw[noarrow,  shorten <=0mm] (q_out) to (q_out_anchor);
\draw[arrowstyle,  shorten <=0mm] (q_out_anchor) to (cross);


\node[below=20mm of q_net, item_node_green] (ras_tex){Reduced \\ Action Space $\mathcal{A}_r$ \\ $1870 \times 6$};

\node[operation_rec, right=0.5cm of ras_tex] (one_hot)   {one hot \\ encode};

\draw[arrowstyle] (ras_tex) to (one_hot);
\node[right=5mm of one_hot, item_node_green] (hot_enc_ras_tex){hot\\ encoded \\ $\mathcal{A}_r$ \\$1870 \times 37$};
\draw[arrowstyle] (one_hot) to (hot_enc_ras_tex);

\coordinate (space_out_midway) at ($(cross) + (0,-1.65)$);
\draw[noarrow] (hot_enc_ras_tex) to (space_out_midway) ;
\draw[arrowstyle, shorten <=0mm] (space_out_midway) to node[textstyle, right] { Transpose} (cross);

\node[textstyle] [right=0.5cm of cross] (q_out_final) {$Q(s,\cdot)$};
\draw[arrowstyle] (cross) to (q_out_final);
\end{tikzpicture}

\caption{Q-value calculation using linear Q decomposition. The critic outputs Q-values for each action bin, where $Q(s,a_{i,j})$ represents the Q-value of the $j$-th bin of the $i$-th action dimension. The one-hot encoded $\mathcal{A}_r$ masks all but one bin per action dimension before linearly combining them to compute the final Q-value for a specific action combination. The output, $Q(s, \cdot)$, has shape $batch\_size \times |\mathcal{A}_r|$. The argmax operator can be applied along the second dimension to select the best action combination for state $s$.}
\label{fig:factorized_q_function}
\end{figure}

\section{Action space reduction}
\begin{figure}[H]
    \centering
\begin{tikzpicture}[
squarednodered/.style={rectangle, draw=red!60, fill=red!5, thick, minimum size=1mm, font=\scriptsize},
squarednodegreen/.style={rectangle, draw=green!60, fill=green!5, thick, minimum size=1mm, font=\scriptsize},
arrowstyle/.style={->, thick, black!70!black, shorten >=1mm, shorten <=1mm},
textstyle/.style={font=\scriptsize, align=center} 
]
\node[textstyle] (dataset_text) {Dataset};
\node[right=15mm of dataset_text, textstyle] (as_text){Action Space $\mathcal{A}$};
\node[right=10mm of as_text, textstyle] (ras_tex){Reduced \\ Action Space $\mathcal{A}_r$};
\node[textstyle, below= 0.5mm of as_text] (a1_text){$a_1$};
\node[squarednodered] (a10) [below= 0.1mm of a1_text] {0};
\node[squarednodered] (a00) [left= 0.5mm of a10] {0};
\node[squarednodered] (a20) [right=0.5mm of a10]  {0};

\node[squarednodegreen] (a01) [below= 0.5mm of a00] {0};
\node[squarednodegreen] (a11) [right= 0.5mm of a01] {0};
\node[squarednodegreen] (a21) [right=0.5mm of a11]  {1};

\node[squarednodered] (a02) [below= 0.5mm of a01] {0};
\node[squarednodered] (a12) [right= 0.5mm of a02] {1};
\node[squarednodered] (a22) [right= 0.5mm of a12] {0};

\node[squarednodegreen] (a03)  [below= 0.5mm of a02] {0};
\node[squarednodegreen] (a13)  [right= 0.5mm of a03] {1};
\node[squarednodegreen]  (a23)  [right= 0.5mm of a13] {1};

\node[squarednodered]  (a04) [below= 0.5mm of a03]  {1};
\node[squarednodered]  (a14) [right= 0.5mm of a04]  {0};
\node[squarednodered]  (a24) [right= 0.5mm of a14]  {1};

\node[squarednodered] (a05) [below= 0.5mm of a04] {1};
\node[squarednodered] (a15) [right= 0.5mm of a05] {0};
\node[squarednodered] (a25) [right= 0.5mm of a15] {1};

\node[squarednodegreen] (a06) [below= 0.5mm of a05] {1};
\node[squarednodegreen] (a16) [right= 0.5mm of a06] {1};
\node[squarednodegreen] (a26) [right= 0.5mm of a16] {0};

\node[squarednodered] (a07)  [below= 0.5mm of a06] {1};
\node[squarednodered]  (a17) [right= 0.5mm of a07] {1};
\node[squarednodered]  (a27)  [right=0.5mm of a17] {1};

\node[textstyle, above= 0.1mm of a00] (a0_text) {$a_0$};
\node[textstyle, above= 0.1mm of a20] (a2_text) {$a_2$};

\node[inner sep=0pt, scale=0.3] [below= 10mm of dataset_text] (dataset) 
    {\includegraphics[width=.25\textwidth]{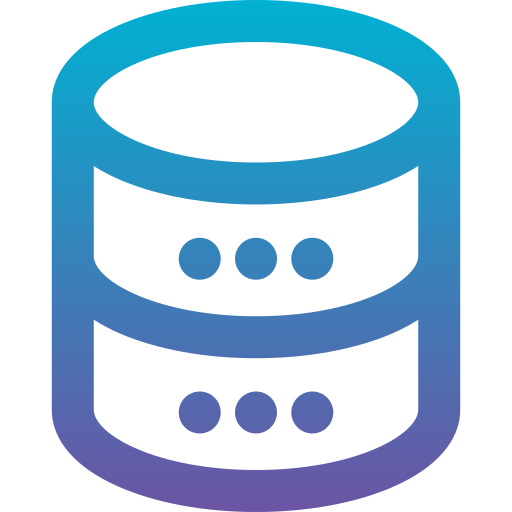}};


\node[textstyle, below= 20mm of ras_tex] (b1_text) {$a_1$};


\node[squarednodegreen]      (b10)                       [below= 0.1mm of b1_text]       {0};
\node[squarednodegreen]      (b00)                       [left= 0.5mm of b10]   {0};
\node[squarednodegreen]      (b20)                       [right=0.5mm of b10]       {1};

\node[squarednodegreen]      (b01)                       [below= 0.5mm of b00]   {0};
\node[squarednodegreen]      (b11)                       [right= 0.5mm of b01]       {1};
\node[squarednodegreen]      (b21)                       [right=0.5mm of b11]       {1};

\node[squarednodegreen]      (b02)                       [below= 0.5mm of b01]   {1};
\node[squarednodegreen]      (b12)                       [right= 0.5mm of b02]       {1};
\node[squarednodegreen]      (b22)                       [right= 0.5mm of b12]       {0};

\node[textstyle, above= 0.1mm of b00] (b0_text) {$a_0$};
\node[textstyle, above= 0.1mm of b20] (b2_text) {$a_2$};

\draw[arrowstyle, out=30, in=150] (a21) to (b00);
\draw[arrowstyle, out=-30, in=170] (a23) to (b01);
\draw[arrowstyle, out=-30, in=-120] (a26) to (b02);

\draw[arrowstyle, out=30, in=150] (dataset) to (a01);
\draw[arrowstyle, out=-60, in=-150] (dataset) to (a03);
\draw[arrowstyle, out=-90, in=-150] (dataset) to (a06);

\end{tikzpicture}

\caption{This example illustrates a 3-dimensional discrete action space, where each dimension has 2 possible values, and each distinct action is represented as $a = [a_0, a_1, a_2]$. The action space is constrained to include only combinations present in the dataset (shown in green), excluding all other combinations (shown in red).}
\label{fig:action_space_reduction}
\end{figure}

\end{document}